\documentclass[journal]{ieeetran}

\usepackage{cite}
\usepackage{amsmath}
\usepackage{algorithmic}
\usepackage{array}

\usepackage{stfloats}
\usepackage{url}
\usepackage{graphicx}
\usepackage{bm}
\usepackage[mathlines]{lineno}
\usepackage{amsmath, amsfonts, amsbsy, amssymb, amsthm}
\usepackage[utf8]{inputenc}
\usepackage[T1]{fontenc}
\usepackage{mathptmx}

\usepackage{array,url,kantlipsum}
\usepackage{multicol,lipsum,xparse}

\usepackage{url}
\usepackage{setspace}
\usepackage[table]{xcolor}
\usepackage{verbatim}
\usepackage[export]{adjustbox}
\usepackage{subcaption}
\usepackage{accents} 
\usepackage{float}
\usepackage{hyperref}
\usepackage{algorithm}
\usepackage[utf8]{inputenc}
\usepackage{multirow}
\usepackage{textcomp}

\title{Accelerating Matrix Diagonalization through Decision Transformers with Epsilon-Greedy Optimization}
\author {Kshitij Bhatta $^{1,3,*}$,
Geigh Zollicoffer $^{2,4}$, Manish Bhattarai$^{4}$, Phil Romero$^3$, Christian F. A. Negre$^4$,Anders M. N. Niklasson$^{4}$ and Adetokunbo Adedoyin$^{5}$

\thanks {This manuscript has been approved for unlimited release and has been assigned LA-UR-24-25918. This
work was supported by the Laboratory Directed Research and Development program of Los Alamos
National Laboratory (LANL) under project number 20220428ER. Kshitij Bhatta and Geigh Zollicoffer were supported by the LANL Applied Machine Learning Summer Research Fellowship (2023) and NSF MSGI Program (2023) respectively. This research used resources provided by the LANL Institutional Computing Program. LANL
is operated by Triad National Security, LLC, for the National Nuclear Security Administration of U.S.12Department of Energy (Contract No. 89233218CNA000001). Additionally, we thank the CCS-7 group, Darwin cluster, and Institutional Computing (IC) at LANL for computational resources. Darwin is funded by the Computational Systems and Software Environments (CSSE) subprogram of LANL’s ASC program (NNSA/DOE).
}
\thanks{$^{1}$ Department of Mechanical and Aerospace Engineering, University of Virginia, Charlottesville, VA, 22904, USA.}
\thanks{$^{2}$ School of Mathematics, Georgia Institute of Technology, Atlanta, GA ,30332, USA}
\thanks{$^{3}$ High Performance Computing division, Los Alamos National Laboratory, NM, 87545, USA.}
\thanks{$^{4}$ Theoretical division, Los Alamos National Laboratory, NM, 87545, USA.}
\thanks{$^{5}$ Computational Science Division, Los Alamos National Laboratory, NM, 87545, USA.}
\thanks {$^*$Corresponding author, Email: qpy8hh@virginia.edu}
}
\begin{document}
\maketitle

\begin{abstract}    
This paper introduces a novel framework for matrix diagonalization, recasting it as a sequential decision-making problem and applying the power of Decision Transformers (DTs).  Our approach determines optimal pivot selection during diagonalization with the Jacobi algorithm, leading to significant speedups compared to the traditional max-element Jacobi method. To bolster robustness, we integrate an epsilon-greedy strategy, enabling success in scenarios where deterministic approaches fail. This work demonstrates the effectiveness of DTs in complex computational tasks and highlights the potential of reimagining mathematical operations through a machine learning lens.  Furthermore, we establish the generalizability of our method by using transfer learning to diagonalize matrices of smaller sizes than those trained.
%\n
 % \noindent\textbf{Keywords:} Eigensolver, Jacobi rotations, Reinforcement Learning, Decision Transformer, AlphaZero
\end{abstract}
\section{Introduction} 
\label{intro}
Eigensolvers play a pivotal role in contemporary scientific computing, addressing a vast array of complex problems across numerous disciplines. These algorithms are designed to efficiently compute eigenvalues and corresponding eigenvectors of matrices, providing valuable insights into the underlying structures and properties of diverse systems \cite{Golub2013, Parlett1998}. The significance of eigensolvers %stems from their ability to identify dominant modes of behavior in a wide range of applications, 
ranges from applications in quantum mechanics and computational chemistry \cite{Szabo1996, Leach2001} to data analysis and machine learning \cite{Jolliffe2016, Goodfellow2016}. Eigenvalue problems arise ubiquitously in scientific and engineering fields when seeking solutions to critical issues such as mode stability, dynamic behavior, and quantum energy levels \cite{Kantorovich1982, Saad2011}. The study of eigenvalues and eigenvectors is vital in understanding the stability and growth of dynamical systems, natural frequencies in mechanical structures, the stability of quantum systems, and the principal components in data analysis \cite{Lanczos1950, Horn2012}.
%In essence, 

Eigensolvers seek to solve the fundamental equation $Av = \lambda v$, where $A$ is a given square matrix, $\lambda$ represents the eigenvalue, and $v$ corresponds to the eigenvector associated with the eigenvalue. Despite the seemingly straightforward formulation of the problem, the efficient computation of eigenvalues and eigenvectors becomes increasingly challenging as the size of the matrix grows, necessitating the development of sophisticated numerical algorithms. The selection of an appropriate eigensolver depends on the matrix characteristics, such as size, sparsity, and structure. Numerous methods have been proposed over the years, each tailored to exploit specific matrix properties and provide computational advantages \cite{Golub2013, Trefethen1997}. These methods encompass various techniques, including iterative methods \cite{Saad2011, Stewart2001}, direct methods \cite{Trefethen1997, Golub2013}, and subspace iteration techniques \cite{Lehoucq1996, Parlett1998}.
With the advent of high-performance computing and the surge in data-centric applications, the demand for efficient eigensolvers has never been higher. This demand is further amplified by the increasing complexity of the problems at hand, many of which involve large-scale matrices with millions, or even billions, of elements. The sheer size and intricacy of these matrices present formidable computational challenges that are not easily addressed by traditional methods. Recent developments in artificial intelligence and machine learning offer a fresh perspective on this age-old problem. Deep learning models, with their ability to capture intricate patterns and relationships, have shown promise in emulating and, in some cases, outperforming classical algorithms \cite{romero2023matrix}. 
One such paradigm-shifting approach is the Decision Transformer (DT)\cite{chen2021decision}. DT is a framework which utlizies the transformer architecture to solve sequential decision making problems. Harnessing the capabilities of DTs, we can effectively transform the task of matrix diagonalization into a sequence-to-sequence learning problem. The resulting innovative method combines the strengths of transformer models, known for their success in natural language processing, with the intricacies of eigensolvers, presenting a unique blend of computational and algorithmic strategies. In this paper, we delve deep into the matrix diagonalization challenges and present an enhanced Decision Transformer model fortified with an epsilon-greedy strategy, ensuring robustness and efficiency in matrix diagonalization tasks. Through evaluations and comparisons with standard heuritstics, we demonstrate the potential of melding deep learning techniques with traditional numerical algorithms, establishing a real time eigen-solver. 
\section{Related work}
\label{rel work}

Matrix diagonalization is a well-studied problem in computational linear algebra. Various algorithms like QR decomposition \cite{anderson1992generalized}, Jacobi's method \cite{cardoso1996jacobi}, and more recently, the AI accelerated Fast Eigenvalue (FastEigen) algorithm based on the Jacobi method \cite{romero2023matrix} have shown promising results in different contexts. While traditional methods focus on computational and algorithmic aspects, recent works like FastEigen have begun to consider the integration of machine learning to solve this challenging problem.  The Decision Transformer aligns closely with recent advances in RL which have demonstrated the ability to learn optimal policies for a wide range of tasks \cite{chen2021decision}. Model-free algorithms, in particular, are attractive for problems where the environmental dynamics are unknown or too complex to model \cite{degris2012model}. The adaptation of transformer models into the RL paradigm is an emerging trend. Earlier works have shown the utility of transformer architectures for natural language tasks \cite{vaswani2017attention}. More recent works have extended this model to RL tasks, including but not limited to robotic control \cite{siebenborn2022crucial},game playing \cite{lee2022multi} and few-shot policy generalization \cite{xu2022prompting}.%, and matrix diagonalization~\cite{romero2023matrix}.
The epsilon-greedy strategy is a widely used exploration technique in RL \cite{tokic2010adaptive}. By introducing a random element in decision-making, it balances exploration and exploitation, and has been applied in various domains, including multi-armed bandits \cite{collier2018deep} and deep RL \cite{silver2018general}. 
The utilization of sequence-to-sequence models in scientific computing tasks is relatively novel but has demonstrated promising results \cite{sutskever2014sequence}. Employing a DT to effectively map the problem of matrix diagonalization into a sequence-to-sequence task represents a notable contribution in this domain.

%Therefore, this paper introduces a scalable and robust approach for matrix diagonalization employing Decision Transformers and epsilon-greedy exploration. Extensive evaluations are presented to showcase the superiority of our method over existing algorithms, demonstrating improved computational efficiency and robustness to real-world variations. Additionally, transfer learning is showcased through the application of a trained model for large matrix sizes, facilitating direct evaluation on smaller matrices.
\section{Methods}
\label{method}
We formulate the matrix diagonalzation problem as a fully observable Markov Decision Process (MDP) defined by the tuple $(\mathcal{S}, \mathcal{A}, P, R,\gamma)$ characterized by states ($\mathcal{S}$), actions ($\mathcal{A}$), transition probabilities ($\mathcal{P}$), rewards ($\mathcal{R}$), and a discount factor ($\gamma$). In this framework, states  represent all possible matrix configurations, and actions  encompass transformations towards diagonalization. The transition function $P(s', s, a)$ defines the likelihood of moving from one matrix configuration $s$ to another $s'$ given an action $a$, encapsulating the problem's dynamics. Rewards $R$ are allocated for actions $a$ that effectively advance a matrix $s$ towards its diagonal form $s'$, with the immediate reward function $R(s, a, s')$ encouraging efficiency and penalizing less effective steps. The discount factor $\gamma$ balances immediate rewards against long-term outcomes, guiding the strategy towards optimal diagonalization. The aim is to discover a policy $\pi$ that maximizes rewards across diagonalization episodes, employing model-free Reinforcement Learning algorithms as a versatile solution for navigating the complexities of matrix transformations without requiring explicit knowledge of the environment's dynamics. This method provides a structured yet flexible approach to solving the matrix diagonalization problem, adapting to various matrix characteristics and optimization criteria.

\subsection{MDP Framework: The Jacobi Rotation Game}
\label{J-game}
The Jacobi algorithm is a classic iterative method used to diagonalize symmetric matrices. It operates by iteratively applying rotations to the matrix to eliminate off-diagonal elements, eventually yielding a diagonal matrix \cite{cardoso1996jacobi}. A commonly used variation of the Jacobi algorithm is the max-element Jacobi algorithm (\textit{Max-elem}) where each iteration involves identifying the largest off-diagonal element and performing a rotation to zero it out \cite{Golub2013}. This process continues until the off-diagonal elements are sufficiently small, indicating convergence to the diagonal form. Given the suitability of the Jacobi algorithm for formulation into a MDP, we opt for this method to demonstrate the efficacy of DTs as eigenvalue solvers. Despite its limitations, notably its applicability only to symmetric matrices, the Jacobi algorithm serves as a robust proof of concept.
The Jacobi rotation game is formulated using the following state, action and reward definitions:

\begin{enumerate}
    \item \textbf{State:} The state of the MDP is the matrix that is to be diagonalized. Normalizing the state using mean and standard deviation is not possible for this problem since doing so changes the eigenvalues of the system. Thus, different normalization strategies have to be considered. Particularly, we explore two strategies.
    \begin{itemize}
        \item No Normalization: The first strategy is to skip normalization all together. 
        \item Scaling by max element: The second strategy is to scale the matrix using the maximum element in the matrix. This makes it such that all elements in the matrix have a value of $1$ or less than $1$. Dividing by a scalar value just scales the value of the eigenvalues by the same number and does not change the eigenvectors ~\cite{PHILLIPS1996299}. 
    \end{itemize}

\item \textbf{Action:} The action within this MDP framework involves selecting one of the off-diagonal elements of the matrix to apply a Jacobi rotation. For symmetric matrices, the diagonalization process can be streamlined by focusing solely on the upper diagonal elements. This simplification arises because the pivots required to diagonalize the upper diagonal block are the transpose of those needed for the lower diagonal block.  Specifically, the action is denoted by a pair of indices $(i, j)$, where $i < j$, representing the pivot elements for the rotation. The chosen action aims to reduce the magnitude of the off-diagonal elements, thereby pushing the matrix closer to a diagonal form with each step. The action space, therefore, consists of $\frac{n(n-1)}{2}$ possible actions for an $n \times n$ matrix, reflecting the number of off-diagonal elements. For ease of reference, the pivot elements are numbered from $0$ to $\frac{n(n-1)}{2}-1$, starting from the top left off-diagonal element and moving rightwards to the bottom right off-diagonal element.

\item \textbf{Reward:} The reward function is designed to quantify the effectiveness of an action towards achieving diagonalization. A straightforward approach is to measure the increase in the number of zeros in the upper off-diagonal elements of the matrix (due to symmetric nature of the matrix). But this does not punish the agent for taking longer to diagonalize nor does it reward the agent for achieving a diagonalized state. Thus two constants to do the same are added
\begin{equation}\label{eq:reward}
    R(s,a,s') = N_{a}-N_{b}-c_{step}+c_{diag}^*
\end{equation}
%Frobenius norm of the upper off-diagonal elements of the matrix (due to symmetric nature of the matrix). Thus, the reward can be defined as the negative change in the sum of squares of the off-diagonal elements before and after an action:

% \[
% R(s, a, s') = \left(\sum_{i=1}^{n-1} \sum_{j=i+1}^{n} |a_{ij}|^2 - \sum_{i=1}^{n-1} \sum_{j=i+1}^{n} |a'_{ij}|^2\right)
% \]
%where $a_{ij}$ and $a'_{ij}$ are the off-diagonal elements of the matrix before and after applying the Jacobi rotation, respectively. This reward function motivates actions that effectively reduce the magnitude of off-diagonal elements, steering the matrix towards diagonalization.
where $N_b$ and $N_a$ represent the number of zeros in the off-diagonal elements of the matrix before and after applying the Jacobi rotation, respectively. $c_{step}$ denotes the constant penalty incurred for each step the matrix remains undiagonalized, while $c_{diag}$ serves as a constant reward upon achieving diagonalization. The superscript $*$ on $c_{diag}$ indicates that this reward component is applicable only when the matrix is fully diagonalized.

\end{enumerate}

\begin{figure}[t!]
    \centering
    \includegraphics[width=0.5\textwidth]{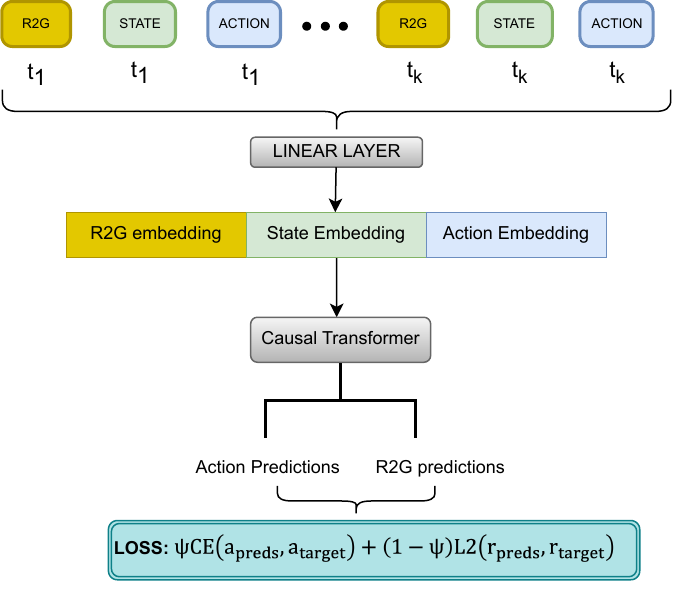}
    \caption{Training architecture of decision transformer.}
    \label{fig:train-arch}
\end{figure}

\subsection{Decision Transformer for matrix diagonalization}
\label{DT}
Decision transformer is a state-of-the-art reinforcement learning framework that conceptualizes sequential decision making problems as conditional sequence modelling problems and leverages the power of causal transformers to autoregressively generate actions that lead to better returns ~\cite{chen2021decision}. The decision transformer takes in "trajectories" as training data and learns the patterns of how taking certain actions in certain states leads to certain returns. Since the model is supposed to generate actions based on future desired returns, rather than past rewards, a new concept of return-to-go is introduced which is the sum of rewards from the current state to the end of the sequence.
\begin{equation}\label{R2G}
    R2G_t=\sum_{t'=t}^T r_t'
\end{equation}
In the context of applying Decision Transformers for matrix diagonalization, the causal transformer architecture emerges as a key component. This methodology utilizes sequences of Returns-to-go (\(R2G\)), states, and actions  denoted as \(rg_t\), \(S_t\), and \(a_t\) respectively, to encapsulate the dynamics of the diagonalization process, as illustrated in Figure~\ref{fig:train-arch}. Initially, each sequence element undergoes a transformation through dedicated linear layers to generate embeddings: \(e_{S_t} = f_{\text{embed}}(S_t)\), \(e_{a_t} = f_{\text{embed}}(a_t)\), and \(e_{rg_t} = f_{\text{embed}}(rg_t)\). These embeddings convert the raw numerical data into a representation within a high-dimensional space allowing for a more nuanced understanding of each element.

The sequences, once embedded, are input into the causal transformer model (\(O_t)\). Contrary to traditional transformers, which permit attention across the full input sequence, the causal transformer imposes a restriction that confines attention to preceding positions only. This mechanism is critical for preserving the temporal order inherent to the matrix diagonalization task, where future actions and rewards cannot logically influence past or current decisions.

The causal transformer then processes these sequences, employing its self-attention mechanism to intricately understand the dependencies between different states, the impact of actions taken, and the expected future rewards. By doing so, it learns to accurately predict the next best action to take for further diagonalization and estimates the associated returns-to-go, effectively guiding the diagonalization process towards completion with optimal efficiency. This comprehensive understanding enables the model to precisely forecast the most advantageous subsequent action (\(\hat{a}_{t+1} = g_{\text{action}}(O_t)\)) and the expected returns-to-go (\(\hat{\text{rg}}_{t+1} = g_{\text{rg}}(O_t)\)) where $g_\text{action}$ and $g_{rg}$ are prediction heads for action and R2G respectively. This steers the diagonalization process towards an efficient and effective resolution. One important consideration is that the decision transformer necessitates prior knowledge of the optimal return for the given problem. This is because the sequence generation is future return conditioned and the model generates actions to try to achieve this return. In this scenario, it's imperative to have an estimate of the optimal number of steps required to diagonalize a matrix. Fortunately, this information can be obtained using either the lower bounds of the steps required by the \textit{Max-element} method or the steps required by the FastEigen~\cite{romero2023matrix} method. Initially, employing a conservative estimate and gradually reducing the number of steps until performance diminishes can be a viable approach.

\subsection{Dataset Construction: Trajectory Generation and Characterization}

Our datasetset consists of trajectories meticulously generated using an environment that performs Jacobi rotations when provided with symmetric matrices and sequence of pivots. Each trajectory, a compilation of tuples (state, action, R2G), encapsulates the journey of matrix diagonalization, portraying the intricacies and decision-making processes involved in reaching a diagonalized state. Here, the \textit{state} reflects the matrix's configuration at any step, the \textit{action} signifies the chosen pivot for diagonalization, and \textit{R2G} quantifies the expected cumulative reward from the current configuration to the desired diagonalized outcome.

\begin{figure*}[t!]
    \centering
    \includegraphics[width=.9\textwidth]{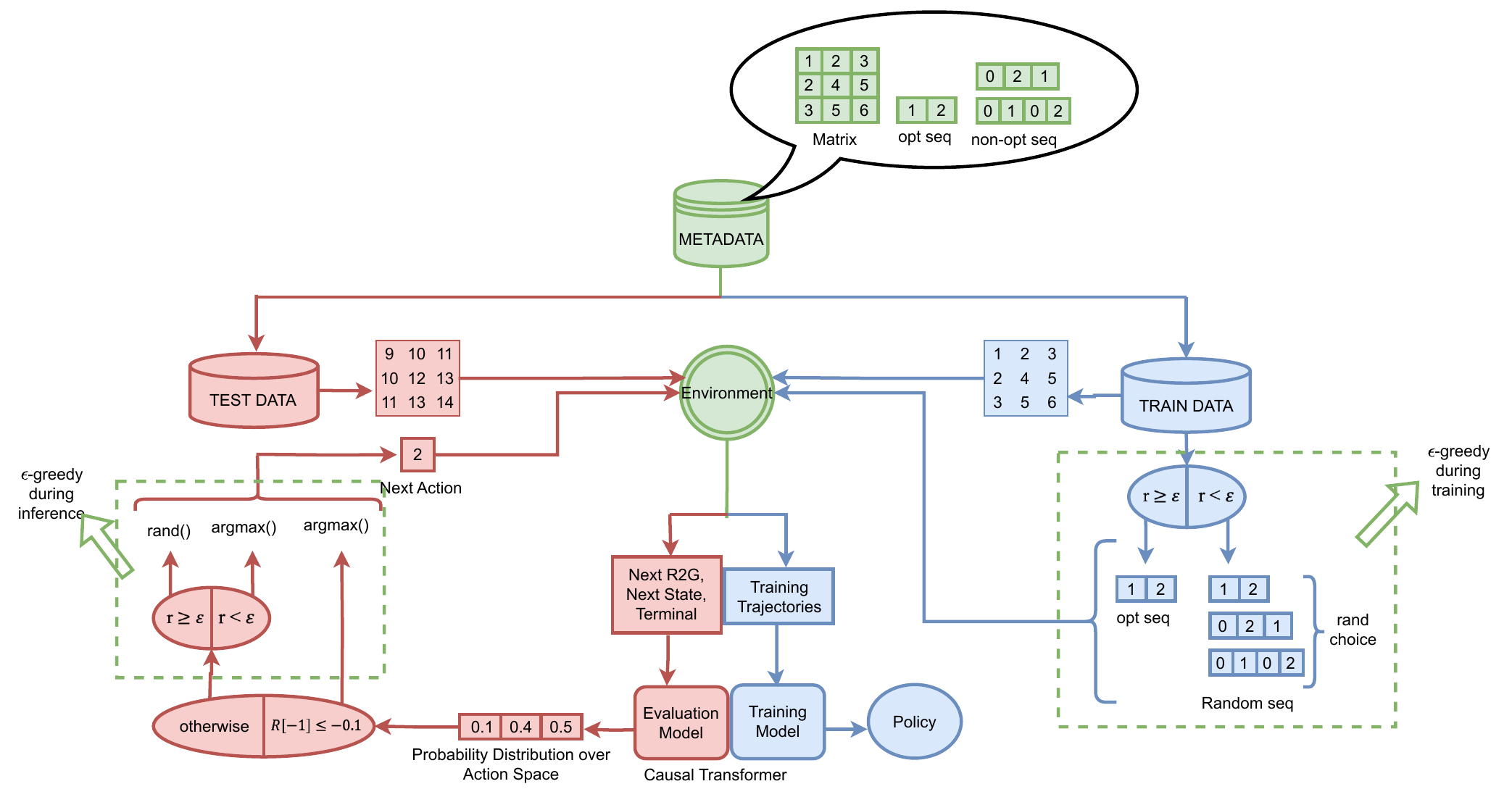}
    \caption{Overall training and inference pipeline. Blue outline represents the training pipeline and red outline represents the evaluation pipeline.}
    \label{fig:train-infer}
\end{figure*}

The dataset is constructed based on metadata sourced from the FastEigen paper~\cite{romero2023matrix}. This metadata consists of Hamiltonian matrices and the sequence of pivots for successful diagonalization. Hamiltonian matrices are a specific type of matrix commonly encountered in molecular dynamic simulations and are characterized by their symmetric nature. The authors of ~\cite{romero2023matrix} utilized the LAMMPS driver for molecular dynamics simulation to conduct NVT simulations at a statistical temperature of 300K. NVT simulation is a type of molecular dynamics simulation where the number of particles (N), volume (V), and temperature (T) are kept constant. This resulted in 1000 $5 \times 5$ symmetric matrices. To introduce greater variability in the matrix elements, sequences were also produced at temperatures of 400K and 500K by the authors. For each of these temperatures, we extracted 1000 matrices, culminating in a total of 3000 matrices. We subdivided these matrices, with 75\% allocated for training and the remaining 25\% reserved for testing.

The sequences in the metadata were generated through the Monte Carlo Tree Search (MCTS) methodology, with over 10,000 playouts for each considered matrix. This approach involves simulating the matrix diagonalization process across a multitude of paths, each delineated by a unique sequence of diagonalization indices. An incremental tree construction method underpins this process, with each node embodying a matrix state post-diagonalization action. The branching from each node represents the array of possible pivot choices for subsequent actions, thus capturing a diverse range of diagonalization paths. We categorized these sequences into optimal and non-optimal groups based on their efficiency in achieving diagonalization. Optimal sequences are those that represent the most streamlined approach, utilizing the fewest possible actions to reach a diagonalized matrix. These sequences are vital for teaching the model the most effective diagonalization strategies. Conversely, non-optimal sequences achieve the same end but via a more circuitous route, entailing a higher number of actions. The inclusion of both sequence types is instrumental in training the model to understand a broad spectrum of potential actions and their outcomes, thus enhancing its ability to generalize across various scenarios.

Using the metadata, trajectories are generated on-the-fly during training by executing jocobi rotations on selected matrices using their respective pivot sequences. Rewards are also allocated by the environment based on the reward definition in section \ref{J-game}. The R2G values are calculated from the rewards using Equation \eqref{R2G}. The resulting dataset comprises a rich variety of trajectories, each a series of (state, action, R2G) tuples leading to the matrix's diagonalization. This diversity, stemming from the MCTS-generated data, ensures that our model is exposed to both optimal and non-optimal diagonalization strategies. Such comprehensive training is pivotal in enabling the causal transformer model to not only predict the most effective actions for matrix diagonalization but also to accurately estimate returns-to-go, thus fostering a nuanced understanding of the diagonalization landscape

% The generation of these sequences is facilitated by an extensive application of MCTS playouts. This approach involves simulating the matrix diagonalization process across a multitude of paths, each delineated by a unique sequence of diagonalization indices. An incremental tree construction method underpins this process, with each node embodying a matrix state post-diagonalization action. The branching from each node represents the array of possible pivot choices for subsequent actions, thus capturing a diverse range of diagonalization paths. As the tree expands, R2G metrics are computed for each path, offering insights into the effectiveness and efficiency of the chosen diagonalization sequence. This metric plays a crucial role in evaluating the desirability of a sequence, encompassing the cumulative expected rewards from a given state to the achievement of a fully diagonalized matrix. .

\subsection{Training and Evaluation Pipeline}
The comprehensive training and evaluation pipeline is depicted in Figure~\ref{fig:train-infer}: training demarcated by a blue outline and evaluation by a red outline. Initially, metadata is partitioned into training and testing subsets. While the training subset encompasses matrices, the optimal sequence, and non-optimal sequences, the testing partition exclusively consists of matrices set for evaluation. 

\subsubsection{Training}
 % For training, the environment recieves a matrix and a sequence of pivots using which it accomplishes the diagonalization task. Thie output of the environment is the trajectory which is a set of matrix states, actions taken (pivots) and the returns-to-go at each state. This generated trajectory then serve as input to our central learning model, the causal transformer. 
During the training phase, the environment is provided with a matrix along with a sequence of pivots necessary for achieving the diagonalization objective. Subsequently, the environment produces a trajectory comprising a collection of matrix states, the actions (pivots) executed, and the corresponding R2G at each state. This resultant trajectory serves as the input data for our core learning model, the causal transformer.

\begin{figure*}[t]
    \centering
    \begin{subfigure}[b]{0.3\textwidth}
        \centering
       \includegraphics[width=0.95\textwidth]{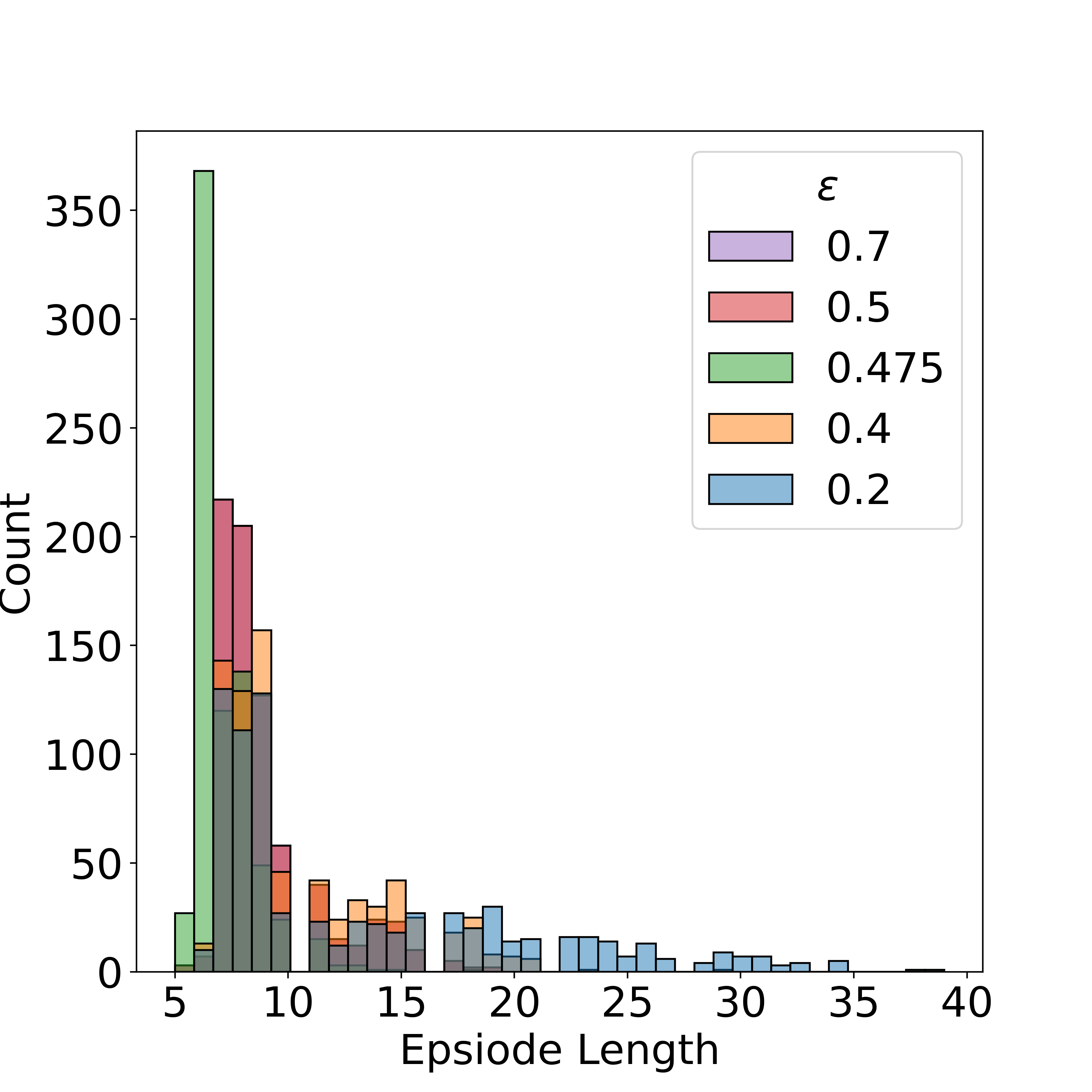}
        \caption{Evaluation for different $\epsilon$ values}
        \label{fig:testing_eps}
    \end{subfigure}
    \hspace{4mm}
    \begin{subfigure}[b]{0.3\textwidth}
        \centering
        \includegraphics[width=0.95\textwidth]{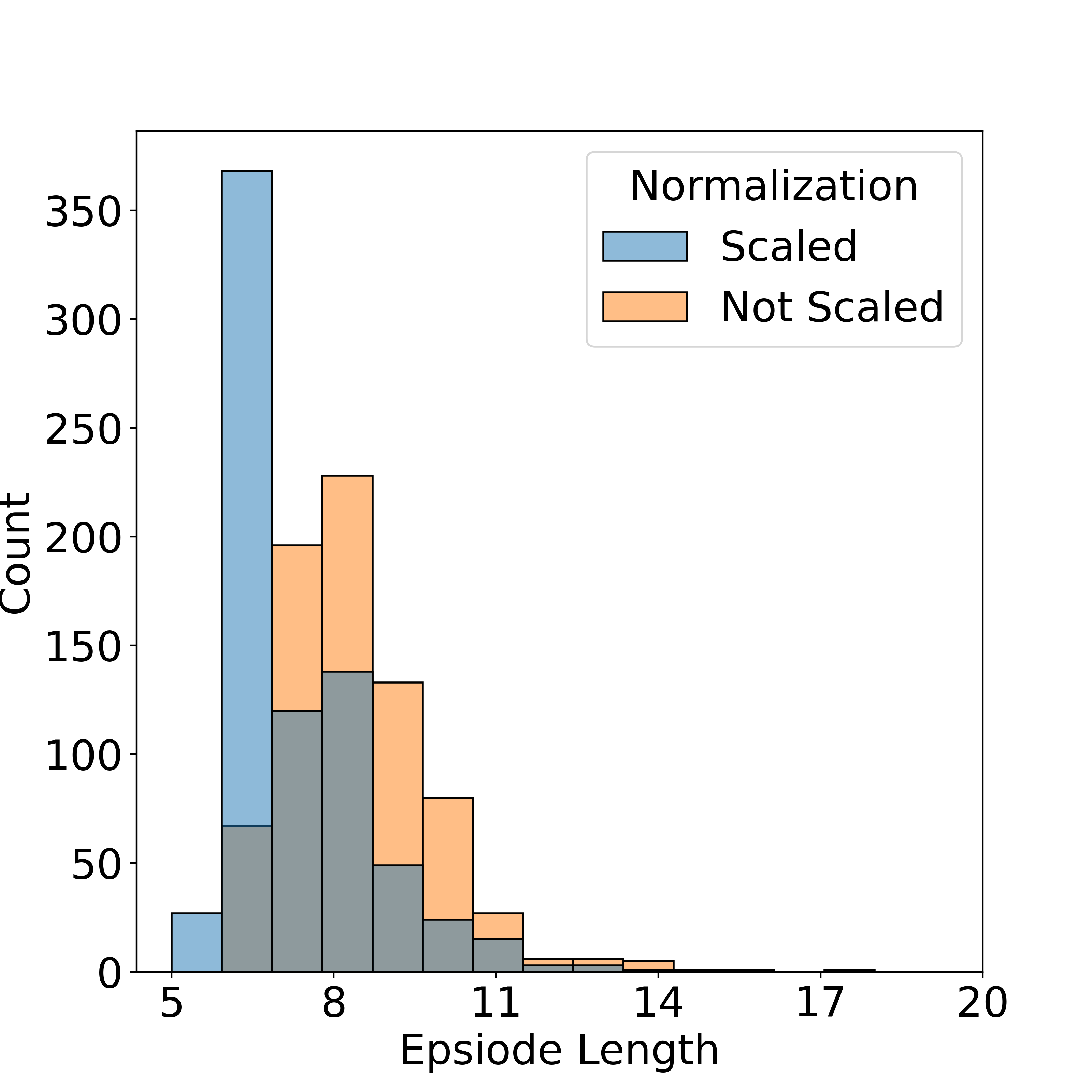}
        \caption{Scaled states vs. Non-scaled states}
        \label{fig:scaled_v_nonscaled}
    \end{subfigure}
    \hspace{4mm}
    \begin{subfigure}[b]{0.3\textwidth}
        \centering
        \includegraphics[width=.95\textwidth]{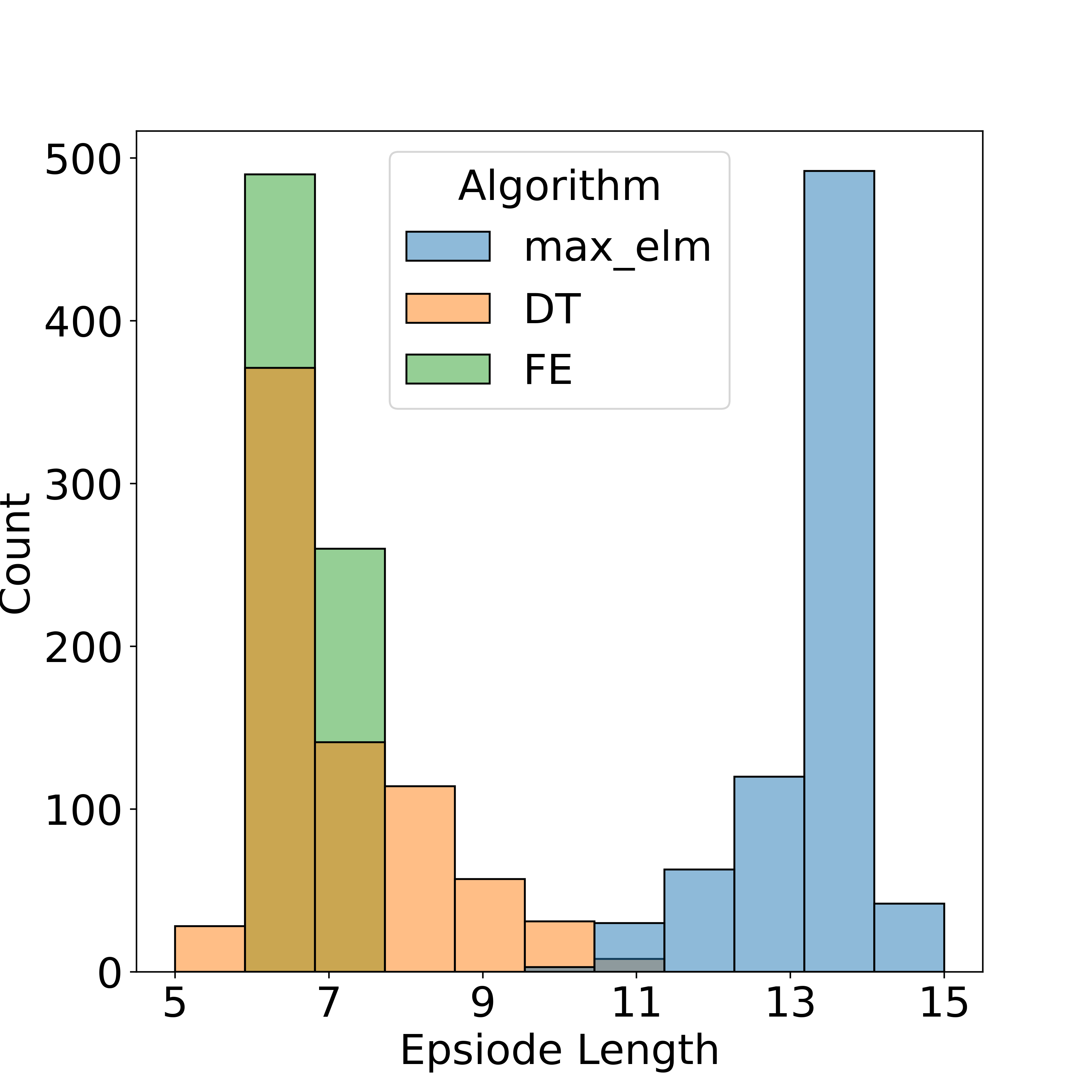}
        \caption{Evaluation using different Algorithms}
        \label{fig:comparison}
    \end{subfigure} 
    \caption{Evaluation Results. Larger episode length implies it takes longer to diagonalize and vice-versa. \emph{Note: The results from FastEigen are derived from the model trained on 500K temperature matrices.}}
    \label{fig:Param_choice}
\end{figure*}

 Given the multiple pivot sequences available for diagonalizing a given matrix, the selection of pivot sequences follows an $\epsilon$-greedy approach. This is a crucial enhancement to the decision transformer's training architecture. Given the Decision Transformer's high susceptibility to its training data, ensuring its exposure to a vast array of states becomes paramount for reinforcing robustness during evaluation. The state space for this problem is expansive: with Hamiltonian matrices for any given temperature having a potentially infinite distribution, introducing the transformer to diverse states, irrespective of their quality, becomes crucial for achieving a comprehensive understanding of the problem landscape. This broad exposure equips the model to better generalize across unseen or less frequent states, thereby enhancing its predictive accuracy and robustness. If the training sequences are represented by the set $\mathcal{Z}$, the epsilon greedy sequence selection proceeds as follows:

\begin{enumerate}
    \item With probability \(\epsilon\), sample uniformly from the available sequences in the training dataset.
         \begin{equation}
        A_{\text{sample}} \sim \text{Uniform} \big(\mathcal{Z}\big)
        \end{equation}
    \item With probability \(1 - \epsilon\), select the optimal sequence from the training dataset: 
     \begin{equation}
        A_{\text{optimal}} = \{x:x\in S|len(x)=min\big(len(\hat{x})\big) \forall \hat{x} \in \mathcal{Z} \}
        \end{equation}
  
\end{enumerate}

The sequence for training is chosen based on these probabilities:
\begin{equation}
    a = 
    \begin{cases} 
        A_{\text{optimal}}, & \text{with probability } 1-\epsilon, \\
        A_{\text{sample}}, & \text{with probability } \epsilon.
    \end{cases}
\end{equation}

The overall training algorithm with inclusion of $\epsilon$-greedy sequence selection is shown in Algorithm~\ref{alg:train-DT}. 

Initially, the trajectories outputted from the environment are embedded into high-dimensional vectors that capture the nuances of each state, action, and R2G value. These embedded sequences are further enriched with positional encodings to account for the sequential nature of the diagonalization process. This enriched data serves as the input for the causal transformer model, designed to handle sequential dependencies by attending only to the current and preceding elements in the sequence as shown in Figure~\ref{fig:train-arch}.

\begin{algorithm}[t]
\caption{Training the DT with $\epsilon$-greedy  for Matrix Diagonalization}
\label{alg:train-DT}
\begin{algorithmic}[1]
\STATE \textbf{Initialize:} DT model parameters $\theta$
\STATE \textbf{Initialize:} Learning rate $\alpha$, regularization parameter $\psi$, and exploration rate $\epsilon$

\FOR{each epoch}
    \FOR{each batch in training set}
        \STATE \textbf{Input:} Matrix states $S$, Set of pivot sequences $\mathcal{Z}$, consisting of optimal sequences $A_{\text{opt}}$ and non-optimal sequences $A_{\text{non-opt}}$
        \STATE \textbf{Epsilon-Greedy Sequence Selection:}
        \STATE \ \ \ \ With probability $1-\epsilon$, select $A_{\text{opt}}$
        \STATE \ \ \ \ With probability $(\epsilon)$, uniformly sample from $\mathcal{S}$
        \STATE \textbf{Forward Pass:}
        \STATE \ \ \ \ Generate trajectories using current matrix states and sequences
        \STATE \ \ \ \ Compute action probabilities and return-to-go (R2G) values using DT
        % \STATE \textbf{Epsilon-Greedy Evaluation}
        % \STATE \ \ \ \ With probability $\epsilon$, update action choice based on $\arg\max$ of policy
        % \STATE \ \ \ \ With probability $(1-\epsilon)$, uniformly sample action from policy
        \STATE \textbf{Loss Calculation:}
        \STATE \ \ \ \ $L = \psi \cdot \text{CE}(a_{\text{preds}}, a_{\text{target}}) + (1-\psi) \cdot L_2(r_{\text{preds}}, r_{\text{target}})$
        \STATE \textbf{Backward Pass:}
        \STATE \ \ \ \ Update DT parameters $\theta$ using gradient descent: $\theta = \theta - \alpha \nabla L$
    \ENDFOR
    \STATE Evaluate DT on test set
\ENDFOR
\end{algorithmic}
\end{algorithm}

%The causal transformer predicts both the action probabilities and the returns-to-go values.
% We use a composite loss function for training, which combines the prediction errors for both actions and R2G. 
% where 9-11 steps are excluded from training in the standard training.

To train the DT, we utilize a composite loss function, \( \mathcal{L} \), which incorporates both the action predictions and reward predictions. The loss function is defined as:

\begin{equation}
    \mathcal{L} = \psi \cdot \text{CE}(a_{\text{preds}}, a_{\text{target}}) + (1 - \psi) \cdot L_2(r_{\text{preds}}, r_{\text{target}})
\end{equation}

where:
\begin{itemize}
    \item \( \psi \) is the regularization parameter balancing the importance between action and reward.
    \item \( \text{CE}(a_{\text{preds}}, a_{\text{target}}) \) is the cross-entropy loss between the predicted actions \( a_{\text{preds}} \) and the target actions \( a_{\text{target}} \).
    \item \( L_2(r_{\text{preds}}, r_{\text{target}}) \) is the mean squared error (MSE) loss between the predicted rewards \( r_{\text{preds}} \) and the target rewards \( r_{\text{target}} \).
\end{itemize}

%For our sequence-to-sequence matrix diagonalization task, the causal transformer's design becomes particularly relevant. The model's inherent inability to access future information in the current step proves invaluable for our problem's sequential nature. 

\begin{figure*}[t]
    \centering
    \begin{subfigure}[b]{0.45\textwidth}
         \centering
        \includegraphics[width=1.1\linewidth]{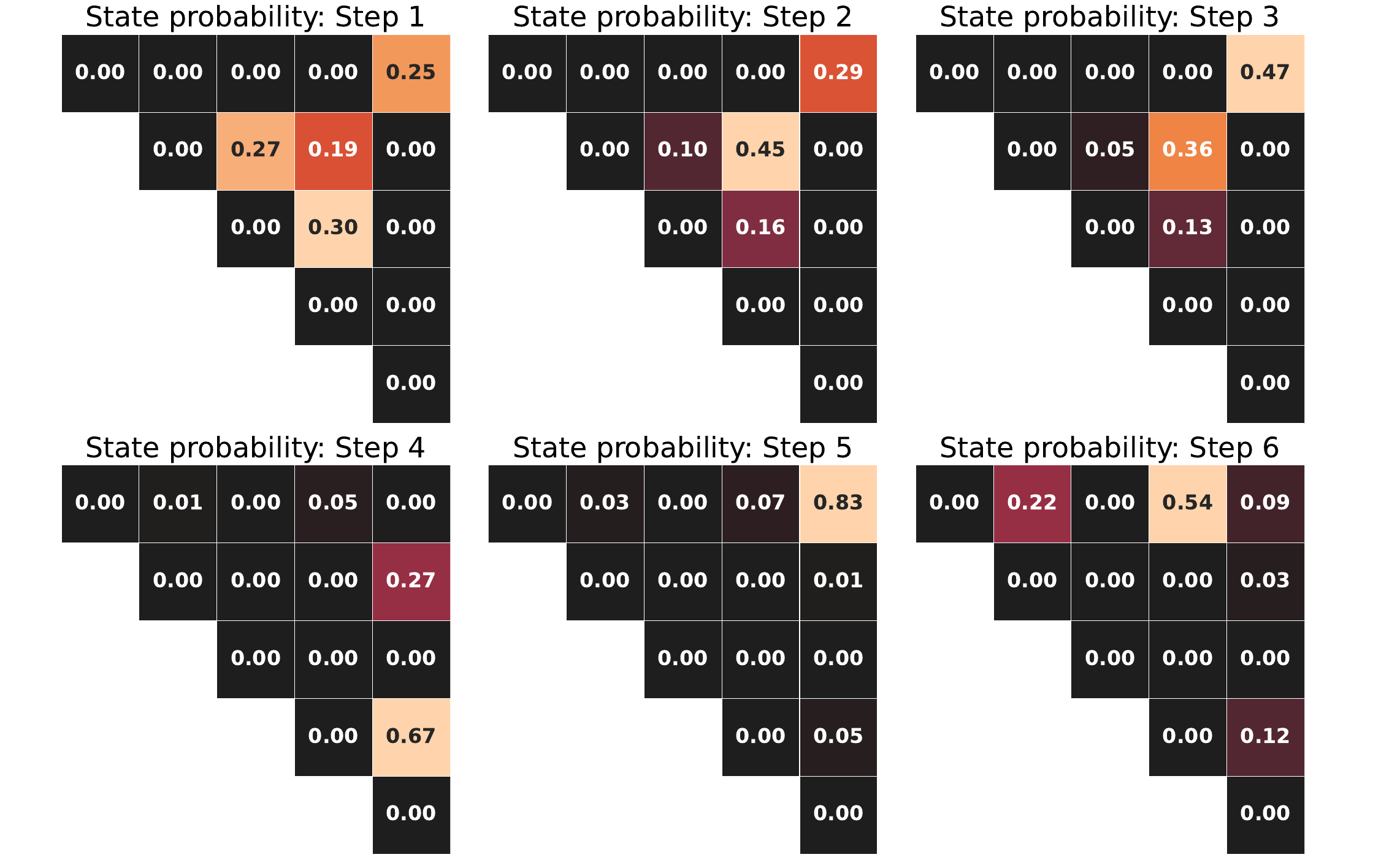}
        \caption{FastEigen action space visualization}
        \label{fig:fasteigenvstatespace_300}
    \end{subfigure}
    \hspace{10mm}
    \begin{subfigure}[b]{0.45\textwidth}
        \centering
        \includegraphics[width=\linewidth]{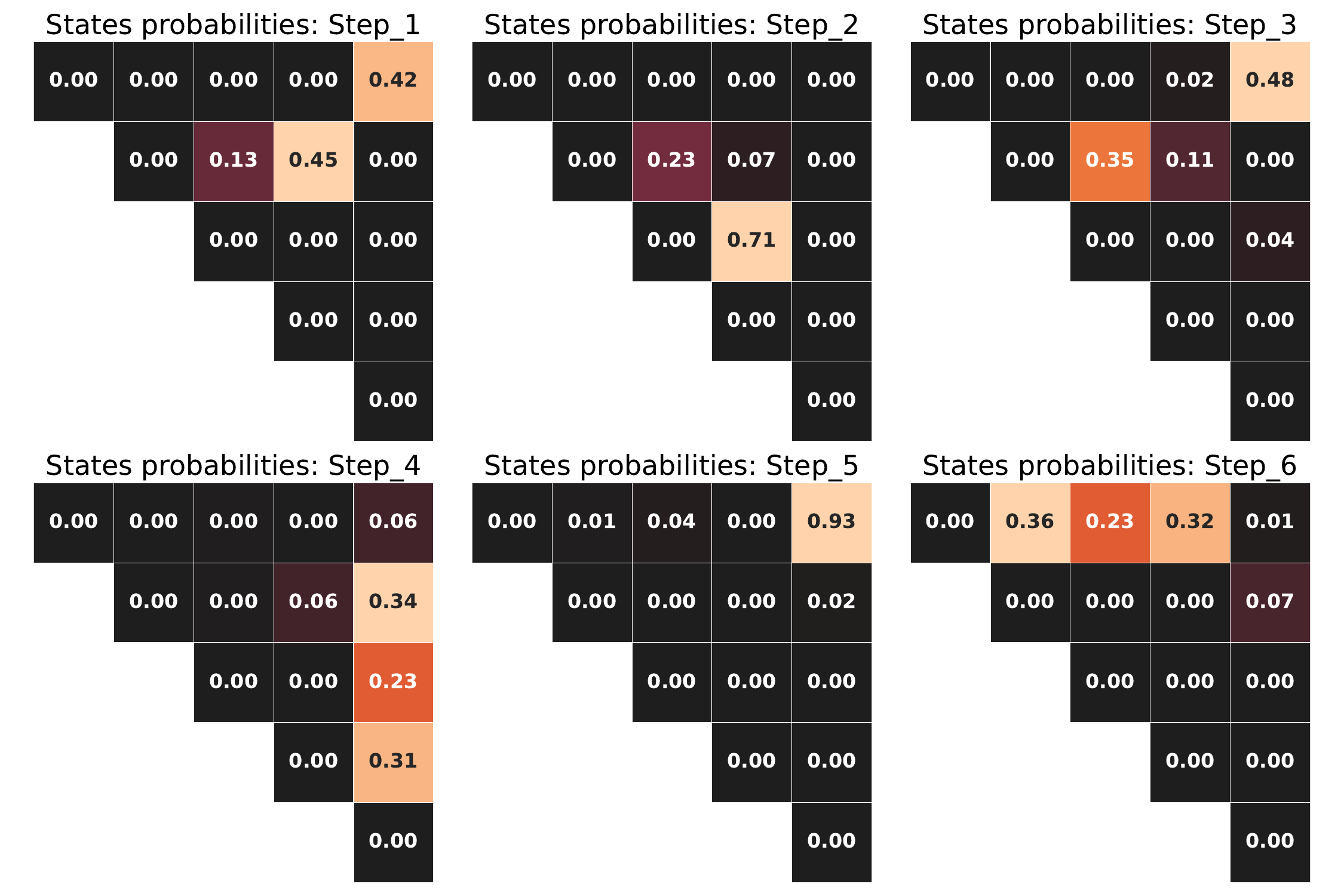}
        \caption{DT action space visualization}
        \label{fig:DT_300}
    \end{subfigure}
    \caption{Probability of pivot points chosen by FastEigen and DT method for  diagonalization. Each matrix represents a step in the diagonalization process, with the numbers within each matrix element indicating the probability of selecting that specific element as the pivot for that step.} 
    %FastEigen is trained on 500K and inferenced on 300K,400K and 500K , DT is trained on 75\% of all temperatures and inferred on 25\%}
    \label{fig:Statespacevis}
\end{figure*}

The training algorithm is shown in Algorithm~\ref{alg:train-DT}. Post-training, the resultant model yields a policy, which essentially denotes the probability distribution over off-diagonal matrix positions for selecting pivots. Every few steps, this policy undergoes evaluation on the test dataset, visualized in Figure~\ref{fig:train-infer} with a red outline. It's imperative to note that the feedback from this evaluation phase doesn't influence the training. Instead, it serves purely as a mechanism to oversee the training's progression.

\subsubsection{Evaluation}

Another innovative amendment made in the Decision Transformer framework in this research is the integration of reward-conditioned $\epsilon$-greedy action selection during the inference phase. Recognizing the limitations of a uniform policy across diverse matrices, particularly in the presence of distributional shifts between training and testing datasets, the $\epsilon$-greedy mechanism emerges as a crucial instrument for directing the matrices towards states that may have been encountered during training. The concept revolves around identifying instances where the current policy fails, indicated by a drop in the reward below a predetermined threshold $r_t$, and employing an $\epsilon$-greedy action to potentially steer the state towards a recognized one. When coupled with $\epsilon$-greedy sequence selection, which exposes the Decision Transformer to a myriad of states, the reward-conditioned $\epsilon$-greedy action selection strategy significantly contributes to achieving satisfactory diagonalization even in matrices that the model has not encountered previously. During inference time, several rollouts are performed and the best performing pivot sequences are extracted.

In the problem of matrix diagonalization for an \( n \times n \) matrix, the policy \(\pi\) has to select one pivot from \( \frac{n(n-1)}{2} \) off-diagonal elements. The policy can be represented as a vector \(\pi \in \mathbb{R}^{\frac{n(n-1)}{2}}\), where \(\pi_i\) is the probability of selecting the \(i\)-th off-diagonal element as the pivot.

Given a reward of $r$, the reward conditioned epsilon-greedy inference proceeds as follows:

\begin{enumerate}
    \item If $r<=r_t$:
    \begin{enumerate}
    \item With probability \(\epsilon\), select the action (pivot) corresponding to the maximum predicted probability in the vector \(\pi\):
    \begin{equation}
        a_{\text{best}} = \text{argmax}_{i=1,2,\ldots,\frac{n(n-1)}{2}} \pi_i
    \end{equation}
    
    \item With probability \(1 - \epsilon\), sample uniformly from the available off-diagonal elements:
    \begin{equation}
        a_{\text{sample}} \sim \text{Uniform} \left( \{1,2,\ldots, \frac{n(n-1)}{2} \} \right)
    \end{equation}
    \end{enumerate}
    \item If $r>r_t$: Select the action corresponding to the maximum predicted probability in the vector \(\pi\).
    
\end{enumerate}

The final action \(a\) is chosen based on these probabilities:
\begin{equation}
    a = \begin{cases}
    \begin{cases} 
        a_{\text{best}}, & \text{with probability } \epsilon, \\
        a_{\text{sample}}, & \text{with probability } 1-\epsilon.
    \end{cases}, \text{ if } r<=r_t \\
    a_{\text{best}}, \hspace{43.5mm} \text{ if } r>r_t 
    \end{cases}
\end{equation}

% The overall training algorithm with inclusion of $\epsilon$-greedy is shown in Algorithm~\ref{alg:train-DT}.
% \begin{figure}
%     \centering
%      \includegraphics[width=.6\textwidth]{Figures_DT/Sequence_Data.png}
%     %\includegraphics[width=2.8in]{Figures_DT/Sequence_Data.png}
%     \caption{Caption}
%     \label{fig:enter-label}
% \end{figure}

\section{Results and discussions}
\label{results}

\subsection{Training and Performance}
\label{Train}

% During the training phase for the Decision Transformer, each matrix was accompanied by the top 100 paths derived from FastEigen. Emphasizing exploration, we set exploration rates at 0.7 for suboptimal sequences and 0.3 for optimal ones. Although this strategy led to inferior performance during the training phase, it significantly enhanced performance during inference. For inference, we tested across various epsilon values as shown in Figure~\ref{fig:testing_eps} and $0.475$ was found be optimal. Furthermore, scaling using the max value was found to have a significant impact in the model's performance. This is shown in Figure~\ref{fig:scaled_v_nonscaled}, where both the models are trained using an epsilon of 0.7 and tested using an epsilon of 0.475.

During the training phase, each matrix was paired with the top 100 paths derived from MCTS playouts, comprising optimal path and sub-optimal paths. The $\epsilon$  for sequence selection was set to $0.7$, initially resulting in inferior performance but notably enhancing performance during inference. The target return for the DT was based on the best performance of the FastEigen algorithm, which was $6$ steps. Thus, utilizing Equation \eqref{eq:reward} with $c_{step}=-0.1$ and $c_{diag}=1.0$, the target return was calculated as $1.4$. The causal transformer consisted of $5$ attention blocks with $8$ heads and a context length of $45$. Training utilized a learning rate of $6.6e^{-5}$, a batch size of $64$, and a dropout probability of $0.1$. An embedding size of $256$ was employed, with a weighting constant $\psi$ of $0.5$ for the composite loss function. Training (gradient) steps totaled $10000$ and evaluation on the test set was done after every $20$ steps using $10$ uniformly sampled matrices to track training progress. The training was performed on four NVIDIA A100 GPUs.

\begin{figure*}[t]% [H] is so declass\'e!
\centering
%\begin{minipage}{0.45\textwidth}
\begin{subfigure}{0.45\textwidth}
\includegraphics[width=.9\linewidth]{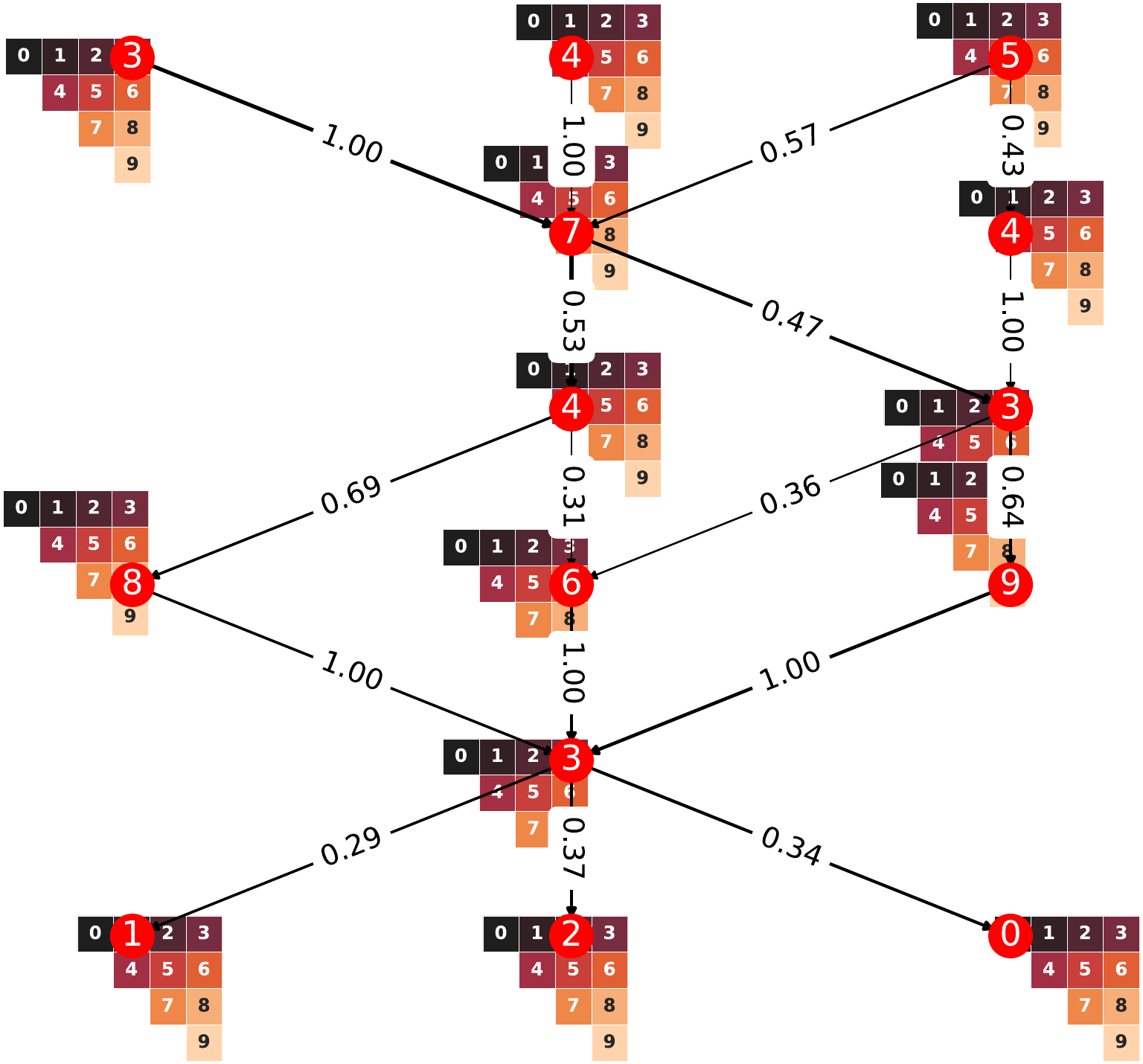}
\caption{}
\label{fig:transisiton_pro_DT}
%\end{minipage}
\end{subfigure}\hfill
%\begin{minipage}{0.45\textwidth}
\begin{subfigure}{0.48\textwidth}
\includegraphics[width=.9\textwidth]{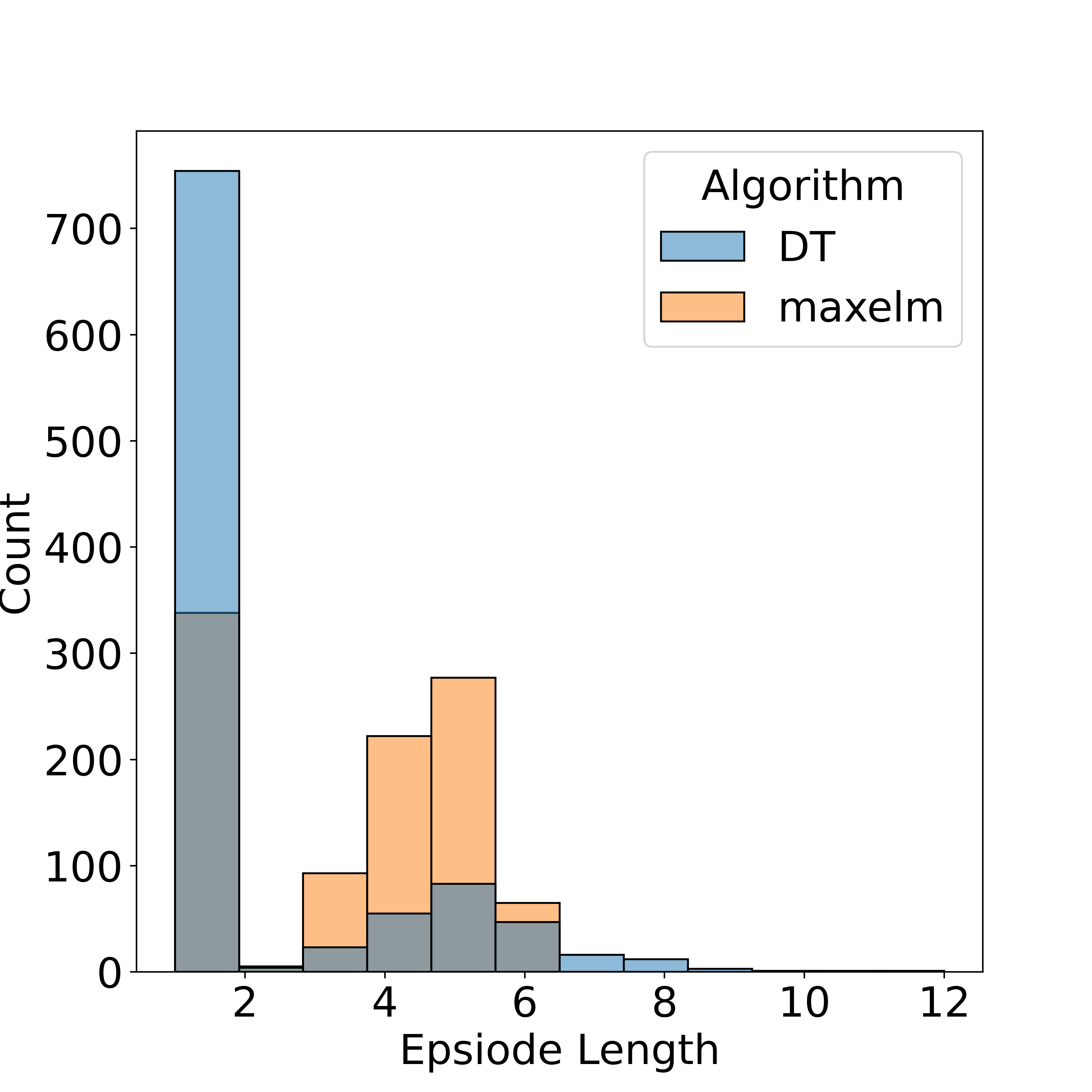}
\caption{}
\label{fig:transfer_learn}
\end{subfigure}
%\end{minipage}
\caption{a) Decision Transformer Action transistion probability. The figure is organized from top to bottom, with each level corresponding to the $i^{th}$ step in the sequence. The numbers enclosed in red circles indicate the potential pivot locations available for use in the $i^{th}$ step. For a detailed explanation of the numbering scheme, please refer to action definition under section \ref{J-game}. Additionally, the numbers displayed on the edges represent the probabilities associated with selecting the pivot indicated by the respective edge in the subsequent step. b) Transfer Learning performance on a 3x3 matrix test set for a DT trained on 5x5 matrices}
\end{figure*}

During inference, $50$ rollouts were performed for each matrix and the best performing pivot sequences were reported. The reward conditioned $\epsilon$ -greedy strategy was set up using a reward threshold of $r_t=-0.1$ which implies taking a step without increasing the number of zeroed out off-diagonal elements. Across various epsilon values tested during inference, $0.475$ emerged as the optimal choice, as depicted in Figure~\ref{fig:testing_eps}. Additionally, scaling using the max value significantly impacted the model's performance, as illustrated in Figure~\ref{fig:scaled_v_nonscaled}, where both models were trained with an $\epsilon$ of $0.7$ and tested with an $\epsilon$ of $0.475$. 

Using this setting, the decision transformer operated within an average path length range of $5-11$. Impressively, it realized an average reduction of $6.692$ steps, a savings of $49.23\%$, in comparison to Max element based diagonalization approach. The performance improvement using our approach compared to other methods can be clearly visualized in Figure~\ref{fig:comparison}. Incorporating the \(\epsilon\)-greedy approach yielded a significant breakthrough for the Decision Transformer. Notably, in certain instances, the algorithm identified more efficient solution paths compared to those found by FastEigen. For example, the Decision Transformer was able to accomplish diagonalization in just $5$ steps, as opposed to the $6$ steps required by the FastEigen algorithm. This finding further underscores the efficacy of our approach, demonstrating the Decision Transformer's ability to not only match but also improve upon existing solutions. 

Although the FastEigen algorithm consistently diagonalizes matrices in $6$ or $7$ steps compared to the broader range of $5$ to $11$ steps for the DT, it's important to note the significant difference in computational requirements. FastEigen relies on $13,000$ rollouts per matrix to achieve its results, whereas our new approach only needs $50$. We can achieve comparable performances to Fasteigen by simply increasing the number of rollouts.  Moreover, the computational time for FastEigen is approximately $5$ minutes per matrix, whereas our method only took $82$ minutes to infer on $750$ matrices, resulting in an inference time of just $0.11$ minutes per matrix resulting a speedup by a factor of $45$. Furthermore, our inference computations were performed on a 14-core 12th generation Intel Core i7 CPU using a Dell XPS 15 laptop compared to FastEigen inference in NVIDIA A100 GPUs.%}

% \begin{figure*}[!ht]
% \centering
% \begin{minipage}{.45\textwidth}
%     \centering
%     \includegraphics[width=\linewidth]{Figures_DT/Results_2.pdf}
%    \subcaption*{a}
%     %\label{fig:fasteigenvstatespace_300}
% \end{minipage}%\hfill
% \begin{minipage}{.45\textwidth}
%     \centering
%     \includegraphics[width=\linewidth]{Figures_DT/Results_3.pdf}
%     \subcaption*{b}
%     %\label{fig:DT_300}
% \end{minipage}
% \caption{Results for Decision Transformers (a) Distribution for average diagonalization path lenth (b) Path length of Maxelem vs Decision Transformer }
% \label{result_img}
% \end{figure*}

% \begin{figure}[t]
%     \centering
%     \includegraphics[width=.4\textwidth]{Figures_DT/testing_eps.png}
%     \caption{Comparison on testing results for different epsilon values}
%     \label{fig:testing_eps}
% \end{figure}

%\subsection{Scalability}
%\label{scale}
\subsection{Understanding Diagonalization Patterns: An Insight into Decision-Making}

Analyzing the probability distribution over the action space for each step taken by the Decision Transformer provides another layer of understanding into the algorithm's functioning. This mapping not only captures the most likely actions to be taken at each step but also offers a nuanced view of how these probabilities evolve over the course of the diagonalization process. For instance, while some actions might have high probability in the initial steps to set the foundation for efficient diagonalization, other actions might gain prominence in later stages, potentially to fine-tune the process or to adapt to specific matrix structures.

To better understand the decision-making process behind our Decision Transformer approach, we analyze the diagonalization patterns and transition probabilities between actions. %Figures~\ref{fig:maxstatespace_300}, 
~\ref{fig:fasteigenvstatespace_300} and ~\ref{fig:DT_300} illustrate the sequential diagonalization steps employed by traditional FastEigen~\cite{romero2023matrix}, and our Decision Transformer model, respectively. Each step represents the series of transformations that ultimately yield a diagonal matrix.

\subsubsection{Pivot Selection Strategy}
In both FastEigen and Decision Transformer models, we observe a strikingly consistent pattern in diagonalization as shown in Figures~\ref{fig:fasteigenvstatespace_300}, and ~\ref{fig:DT_300}. The initial three steps primarily focus on pivot points near the diagonal elements, whereas the later steps shift towards the matrix boundaries. This demonstrates FastEigen's effectiveness in selecting pivotal elements, a strategy well captured by Decision Transformer. Initially targeting diagonal elements proves to be a sound tactic, aiding in the preservation of crucial information during the initial stages of diagonalization. Towards the end, the focus shifts to boundary elements, possibly to exploit the sparse structure of the matrix or to finalize diagonalization by targeting less dominant features.

\subsubsection{Transition Probabilities and Optimal Actions}
Figure~\ref{fig:transisiton_pro_DT} illustrates the transition probabilities between different actions, providing valuable insight into the likelihood of a particular action following another. Constructed from the top $5$ most utilized sequences in the test dataset, this graph captures $64\%$ of the total number of sequences available, indicating the robustness of the algorithm-generated sequences. In the initial steps, three actions (3, 4, and 5) frequently emerge as dominant, setting up the foundation for an efficient diagonalization process. Their prominence in the early stages indicates that they play a crucial role in determining the success of subsequent diagonalization steps. Interestingly, in the fifth step, only a single action (3) appears as the optimal choice, suggesting that this stage is a critical junctures where the correct action is imperative for successful diagonalization. In contrast, the second and the third steps present a bit more flexibility offering two potential actions each (7 and 4 for the second and 4 and 3 for the third), indicating that the choice between these actions depends on the specific state of the system at that moment. The fourth and the sixth steps offer even more flexibility, with three potential actions available for each step. This indicates that the decision-making process here is influenced by additional subtleties in the current state.

\subsubsection{Robustness and Adaptability}
The observed consistency in diagonalization patterns across varying matrices implies that our Decision Transformer approach is both robust and adaptable. This is particularly promising for its potential applications in diverse real-world scenarios, reinforcing the model's value in tackling complex tasks effectively.

The insights extracted from the action space probability distribution can be invaluable for generalizing the algorithm. By identifying consistently high-probability actions or patterns across different matrices and scenarios, we can develop scalable and robust heuristics that may be applied beyond the training set. These heuristics could then serve as a basis for designing more computationally efficient algorithms or for transferring the learned knowledge to tackle different types of matrix diagonalization problems.

% \begin{figure}[t!]
%     \centering
%     \includegraphics[width=.4\linewidth]{Figures_DT/transistion_graph_DT_500_500.pdf}
%     \caption{Decision Transformer Action transistion probability}
%     \label{fig:transisiton_pro_DT}
% \end{figure}

\subsection{Transfer Learning}
In addition to leveraging diagonalization patterns observed from action space distributions for transfer learning, the decision transformer exhibits the capability to directly diagonalize matrices of various sizes without the need for additional training. This is enabled by the symmetric nature of matrices, allowing training on large matrices and subsequent inference on smaller ones with padded outer elements. As depicted in Figure~\ref{fig:transfer_learn}, when the model trained on a 5x5 matrix dataset was tested on a 3x3 matrix dataset, it achieved diagonalization in an average of 1.274 fewer steps than the max-element method, resulting in a performance improvement of $38.72$~$\%$. This indicates that with training on a dataset of sufficiently large matrices, the decision transformer-based eigensolver can directly handle matrices of any size below that without requiring additional training.

% \begin{figure}[h]
%     \centering
%     \includegraphics[width=.4\textwidth]{Figures_DT/output_transfer_3x3.png}
%     \caption{Transfer Learning performance on a 3x3 matrix test set}
%     \label{fig:transfer_learn}
% \end{figure}

% \begin{figure}[t!]
%     \centering
%     \includegraphics[width=.9\linewidth]{Figures_fasteigen/state_diagram_maxelem.pdf}
%     \caption{Maxelemen Action transistion probability}
%     \label{fig:transisiton_prob_max}
% \end{figure}

% \begin{figure}[t!]
%     \centering
%     \includegraphics[width=.9\linewidth]{Figures_fasteigen/state_diagram_fasteigen.pdf}
%     \caption{FastEigen Action transistion probability}
%     \label{fig:transisiton_prob_FE}
% \end{figure}

\section{Conclusion}
\label{conclusion}
In this study, we presented a novel approach to the matrix diagonalization problem by modeling it as a fully observable Markov Decision Process (MDP). We employed a Decision Transformer trained on a unique dataset derived from the FastEigen paper. The DT was designed to learn optimal and sub-optimal sequence paths for matrix diagonalization, incorporating causal attention mechanisms to adhere to the temporal dependencies inherent in the problem. 

One of the key contributions of this work was the introduction of an epsilon-greedy strategy during both the training and inference phases. This inclusion allowed the DT to explore a broader state-action space, enabling the model to generalize better on unseen states, thereby improving robustness. We observed that while the epsilon-greedy approach may lead to suboptimal choices during training, it significantly enhances the performance during inference, occasionally finding better solutions than those reported in previous methods. We also refined the loss function to balance between action prediction and return-to-go prediction, offering the DT more nuanced feedback during the training phase. As a result, our DT not only performed competitively compared to existing state-of-the-art methods, but also exhibited an average reduction of about 49.23\% in the number of steps required for diagonalization, without compromising on accuracy. However, it's crucial to note that the policy's performance could vary depending on the distribution of matrices in the training and testing datasets. This observation underscores the importance of including a stochastic component in the decision-making process, as realized through our epsilon-greedy strategy.

Several avenues for future research can be explored to extend and refine the methodologies presented in this paper. One potential direction is to examine the impact of sampling states based on a probability distribution rather than using a deterministic or epsilon-greedy approach. This heuristic could offer a more nuanced exploration strategy that takes into account the uncertainties in the state transitions and rewards, potentially improving the model's performance in terms of both speed and quality of diagonalization. Another area for improvement is the optimization of the codebase to handle larger sampling sizes efficiently. As the size of the sample space increases, computational demands also escalate. Hence, optimizing the underlying algorithms and data structures will be crucial for scaling up the approach and making it applicable to more complex and large-scale problems.

%Lastly, exploring more sophisticated data generation strategies could also yield improvements. For instance, distributed Monte Carlo Tree Search (MCTS) could be utilized to generate high-quality training data in a more parallelized and efficient manner. This would not only accelerate the training process but could also improve the quality of the policy learned by the Decision Transformer.

\bibliographystyle{unsrt}
\bibliography{main.bib}

\begin{thebibliography}{10}

\bibitem{Golub2013}
Gene~H. Golub and Charles~F. Van~Loan.
\newblock {\em Matrix Computations}.
\newblock Johns Hopkins University Press, 4 edition, 2013.

\bibitem{Parlett1998}
Beresford~N. Parlett.
\newblock {\em The Symmetric Eigenvalue Problem}.
\newblock SIAM, 1998.

\bibitem{Szabo1996}
Attila Szabo and Neil~S. Ostlund.
\newblock {\em Modern Quantum Chemistry: Introduction to Advanced Electronic Structure Theory}.
\newblock Dover Publications, 1996.

\bibitem{Leach2001}
Andrew~R. Leach.
\newblock {\em Molecular Modelling: Principles and Applications}.
\newblock Prentice Hall, 2 edition, 2001.

\bibitem{Jolliffe2016}
Ian~T. Jolliffe.
\newblock {\em Principal Component Analysis}.
\newblock Springer, 2016.

\bibitem{Goodfellow2016}
Ian Goodfellow, Yoshua Bengio, and Aaron Courville.
\newblock {\em Deep Learning}.
\newblock MIT Press, 2016.

\bibitem{Kantorovich1982}
Lev~V. Kantorovich and Vladimir~I. Krylov.
\newblock {\em Approximate Methods of Higher Analysis}.
\newblock Interscience Publishers, 1982.

\bibitem{Saad2011}
Yousef Saad.
\newblock {\em Numerical Methods for Large Eigenvalue Problems}.
\newblock SIAM, 2 edition, 2011.

\bibitem{Lanczos1950}
Cornelius Lanczos.
\newblock An iteration method for the solution of the eigenvalue problem of linear differential and integral operators.
\newblock {\em United States Government Press Office}, 1950.

\bibitem{Horn2012}
Roger~A. Horn and Charles~R. Johnson.
\newblock {\em Matrix Analysis}.
\newblock Cambridge University Press, 2 edition, 2012.

\bibitem{Trefethen1997}
Lloyd~N. Trefethen and David~III Bau.
\newblock {\em Numerical Linear Algebra}.
\newblock SIAM, 1997.
\newblock Discusses a range of numerical methods including direct and iterative methods for solving linear systems, with implications for eigenvalue problems.

\bibitem{Stewart2001}
G.~W. Stewart.
\newblock A krylov–schur algorithm for large eigenproblems.
\newblock {\em SIAM Journal on Matrix Analysis and Applications}, 23(3):601--614, 2001.
\newblock Presents the Krylov-Schur method, an iterative technique for large eigenvalue problems.

\bibitem{Lehoucq1996}
R.~B. Lehoucq and D.~C. Sorensen.
\newblock Deflation techniques for an implicitly restarted arnoldi iteration.
\newblock In {\em SIAM Journal on Matrix Analysis and Applications}, volume~17, pages 789--821. SIAM, 1996.
\newblock Explores deflation techniques in the context of Arnoldi iteration, relevant for subspace iteration methods.

\bibitem{romero2023matrix}
Phil Romero, Manish Bhattarai, Christian~FA Negre, Anders Niklasson, and Adetokunbo Adedoyin.
\newblock Matrix diagonalization as a board game: Teaching an eigensolver the fastest path to solution.
\newblock {\em arXiv preprint arXiv:2306.10075}, 2023.

\bibitem{chen2021decision}
Lili Chen, Kevin Lu, Aravind Rajeswaran, Kimin Lee, Aditya Grover, Misha Laskin, Pieter Abbeel, Aravind Srinivas, and Igor Mordatch.
\newblock Decision transformer: Reinforcement learning via sequence modeling.
\newblock {\em Advances in neural information processing systems}, 34:15084--15097, 2021.

\bibitem{anderson1992generalized}
E~Anderson, Zhaojun Bai, and J~Dongarra.
\newblock Generalized qr factorization and its applications.
\newblock {\em Linear Algebra and its Applications}, 162:243--271, 1992.

\bibitem{cardoso1996jacobi}
Jean-Fran{\c{c}}ois Cardoso and Antoine Souloumiac.
\newblock Jacobi angles for simultaneous diagonalization.
\newblock {\em SIAM journal on matrix analysis and applications}, 17(1):161--164, 1996.

\bibitem{degris2012model}
Thomas Degris, Patrick~M Pilarski, and Richard~S Sutton.
\newblock Model-free reinforcement learning with continuous action in practice.
\newblock In {\em 2012 American Control Conference (ACC)}, pages 2177--2182. IEEE, 2012.

\bibitem{vaswani2017attention}
Ashish Vaswani, Noam Shazeer, Niki Parmar, Jakob Uszkoreit, Llion Jones, Aidan~N Gomez, {\L}ukasz Kaiser, and Illia Polosukhin.
\newblock Attention is all you need.
\newblock {\em Advances in neural information processing systems}, 30, 2017.

\bibitem{siebenborn2022crucial}
Max Siebenborn, Boris Belousov, Junning Huang, and Jan Peters.
\newblock How crucial is transformer in decision transformer?
\newblock {\em arXiv preprint arXiv:2211.14655}, 2022.

\bibitem{lee2022multi}
Kuang-Huei Lee, Ofir Nachum, Mengjiao~Sherry Yang, Lisa Lee, Daniel Freeman, Sergio Guadarrama, Ian Fischer, Winnie Xu, Eric Jang, Henryk Michalewski, et~al.
\newblock Multi-game decision transformers.
\newblock {\em Advances in Neural Information Processing Systems}, 35:27921--27936, 2022.

\bibitem{xu2022prompting}
Mengdi Xu, Yikang Shen, Shun Zhang, Yuchen Lu, Ding Zhao, Joshua Tenenbaum, and Chuang Gan.
\newblock Prompting decision transformer for few-shot policy generalization.
\newblock In {\em international conference on machine learning}, pages 24631--24645. PMLR, 2022.

\bibitem{tokic2010adaptive}
Michel Tokic.
\newblock Adaptive $\varepsilon$-greedy exploration in reinforcement learning based on value differences.
\newblock In {\em Annual Conference on Artificial Intelligence}, pages 203--210. Springer, 2010.

\bibitem{collier2018deep}
Mark Collier and Hector~Urdiales Llorens.
\newblock Deep contextual multi-armed bandits.
\newblock {\em arXiv preprint arXiv:1807.09809}, 2018.

\bibitem{silver2018general}
David Silver, Thomas Hubert, Julian Schrittwieser, Ioannis Antonoglou, Matthew Lai, Arthur Guez, Marc Lanctot, Laurent Sifre, Dharshan Kumaran, Thore Graepel, et~al.
\newblock A general reinforcement learning algorithm that masters chess, shogi, and go through self-play.
\newblock {\em Science}, 362(6419):1140--1144, 2018.

\bibitem{sutskever2014sequence}
Ilya Sutskever, Oriol Vinyals, and Quoc~V Le.
\newblock Sequence to sequence learning with neural networks.
\newblock {\em Advances in neural information processing systems}, 27, 2014.

\bibitem{PHILLIPS1996299}
G.M. PHILLIPS and P.J. TAYLOR.
\newblock Chapter 11 - matrix eigenvalues and eigenvectors.
\newblock In G.M. PHILLIPS and P.J. TAYLOR, editors, {\em Theory and Applications of Numerical Analysis (Second Edition)}, pages 299--322. Academic Press, London, second edition edition, 1996.

\end{thebibliography}

\end{document}